\documentclass{article}
\usepackage{iclr2027_conference,times}

\usepackage{amsmath,amsfonts,bm}

\def\eqref#1{equation~\ref{#1}}

\def\1{\bm{1}}

\DeclareMathAlphabet{\mathsfit}{\encodingdefault}{\sfdefault}{m}{sl}
\SetMathAlphabet{\mathsfit}{bold}{\encodingdefault}{\sfdefault}{bx}{n}

\usepackage[hyperfootnotes=false]{hyperref}
\usepackage{url}
\usepackage{graphicx}
\usepackage{booktabs}
\usepackage{multirow}
\usepackage{colortbl}
\usepackage{tabularx}
\usepackage{placeins}

\newtheorem{theorem}{Theorem}

\title{Inductive Feedback for \\ Mixed-Policy Distillation}

\author{
Amir Moeini\textsuperscript{1,2\,*}\quad
Huaijiang Zhu\textsuperscript{1\,\textdagger}\quad
Daniel Havir\textsuperscript{1}\quad
Shangtong Zhang\textsuperscript{2} \\[2ex]
\normalfont\textsuperscript{1}Alquist Robotics\quad
\textsuperscript{2}University of Virginia
}

\iclrfinalcopy
\hypersetup{
  pdftitle={Inductive Feedback for Mixed-Policy Distillation},
  pdfauthor={Amir Moeini, Huaijiang Zhu, Daniel Havir, Shangtong Zhang}
}

\begin{document}

\maketitle
\fancyhead{}
\renewcommand{\headrulewidth}{0pt}
\begingroup
\renewcommand{\thefootnote}{}
\footnotetext{
\raggedright
\textsuperscript{*}Work started during an internship at Alquist Robotics.
\textsuperscript{\textdagger}Work done when the author was at Alquist Robotics.
Correspondence: Amir Moeini \mbox{\textless\href{mailto:amoeini@virginia.edu}{amoeini@virginia.edu}\textgreater};
Huaijiang Zhu \mbox{\textless\href{mailto:huaijiang.zhu@gmail.com}{huaijiang.zhu@gmail.com}\textgreater};
Daniel Havir \mbox{\textless\href{mailto:dh@alquistrobotics.com}{dh@alquistrobotics.com}\textgreater};
Shangtong Zhang \mbox{\textless\href{mailto:shangtong@virginia.edu}{shangtong@virginia.edu}\textgreater}.
}
\endgroup

\begin{abstract}
Verbal feedback can identify errors and prescribe corrections, providing rich supervision for language-model post-training even when reliable programmatic verifiers are unavailable. Such feedback, often generated by a capable model, can be used to condition the teacher in on-policy distillation, which trains the student to match the teacher's predictions on student-generated rollouts. However, this approach can transfer teacher preferences that the feedback did not motivate, while leaving much of the feedback's guidance unused. We find that both problems come from the standard on-policy distillation objective, specifically the divergence it minimizes and the distribution it uses as its target. Our proposed method addresses both limitations. First, to isolate the information conveyed by the feedback from the teacher's inherent preferences, we treat verbal feedback as evidence for or against the hypothesis that a particular token comes next at a given prefix. We then adopt a probabilistic confirmation framework which uniquely determines an ordering over the vocabulary based on the teacher's predictions before and after it receives feedback. Using a confirmation score consistent with this ordering, we construct a target distribution within a trust region of the student. Second, to learn from guidance that student rollouts can leave unused, we derive a simple shared-rollout estimator of a symmetric divergence between the student and target distributions over rollouts, reusing student and feedback-conditioned teacher rollouts in both directions through importance weighting. Empirical evaluations show that our method outperforms the common on-policy distillation recipe and a recent contrastive variant on knowledge-based and agentic benchmarks.

\end{abstract}

\section{Introduction}
Reinforcement learning (RL) is widely used for post-training language models, improving their reasoning and instruction-following abilities and enabling them to acquire new skills.
A central challenge is assigning credit to individual decisions \citep{lightman2024verify,kazemnejad2025vineppo}.
Methods such as GRPO \citep{shao2024deepseekmath} convert each rollout's scalar reward into a group-normalized advantage and cascade it across the rollout's tokens.
This coarsely reinforces every token in a higher-rewarded rollout, including steps that did not contribute to the outcome.
On-policy distillation (OPD) can provide an alternative way to construct advantages for policy updates \citep{lu2025onpolicydistillation}.
The student is the model being trained, and the teacher is the model whose next-token predictions provide supervision.
OPD constructs token-specific advantages from differences between the teacher's and student's next-token predictions along student-generated rollouts.
These advantages draw on the knowledge encoded in the teacher's distribution to guide the student's decisions.
A useful special case is when the teacher is the student model itself, conditioned on privileged information such as a reference solution or execution feedback \citep{hubotter2026reinforcement}.
The teacher then converts this text-based privileged information into a learning signal in the form of token-level advantages.

One particularly useful form of privileged information is verbal feedback written by an LLM judge \citep{zheng2023judging}. Such feedback is widely used to evaluate language models \citep{liu2023geval, dubois2024alpacaeval} and to refine their rollouts at inference time \citep{madaan2023selfrefine, shinn2023reflexion}.
It is also a valuable source of supervision for post-training.
Where tasks lack reliable programmatic verifiers, a capable judge offers a practical alternative, assessing rollouts against task-specific criteria and communicating its findings through verbal feedback.
Even when outcome rewards are available, verbal feedback still offers a richer basis for learning.
Such feedback can connect specific parts of a rollout to the criteria they satisfy or violate, explaining both where credit belongs and why.
It thereby preserves information lost in reward aggregation and provides step-specific guidance that learning from a single scalar reward must infer across repeated rollouts \citep{xu2026llf}.

These properties make verbal feedback a natural form of privileged information for OPD.
A judge first evaluates the student's rollout and writes feedback; the teacher then conditions on that feedback to provide token-level supervision for the student.
Whereas standard OPD draws its training signal from the teacher's own expertise, here the judge supplies the guidance and the teacher interprets it.
This gives us two additional ways to scale supervision: increasing the judge's model size or its test-time compute.
Because feedback is communicated through text, frontier models can serve as judges even when their token probabilities for the original task are unavailable.
Furthermore, the teacher is conditioned on the student's rollout and performs no further reasoning about the rollout or the criteria, whereas the judge can spend additional test-time compute reasoning about both, improving its feedback and thereby the token-level training signal the teacher provides.
These ways of improving feedback are orthogonal to using a stronger teacher, which can provide better token-level supervision by interpreting that feedback more effectively.
In this paper, we therefore focus on verbal feedback from an LLM judge as the privileged information, aiming to improve the student's behavior without requiring that feedback at inference.
However, applying OPD in this setting can transfer teacher preferences unrelated to the feedback's guidance and leave much of that guidance unused.

First, OPD can transfer too much of the teacher's behavior.
The teacher and student can disagree strongly on tokens that have little to do with either the verbal feedback or solving the task.
For example, stylistic preferences acquired during training can lead them to assign very different probabilities to choices of phrasing and formatting \citep{pan2026rlcsd}.
Such differences can arise even in self-distillation, where the teacher may average the student's parameters over its training history \citep{hubotter2026reinforcement}.
Direct matching turns these discrepancies into training signals, pushing the student to adopt the teacher's habits or limitations along with the behavior we want it to learn.
We want to transfer the change the feedback induces in the teacher's predictions.
To this end, we turn to probabilistic confirmation, treating the feedback as evidence about the hypothesis that a particular token comes next.
The teacher's prediction without feedback supplies the prior probability of that hypothesis, while its prediction conditioned on the feedback supplies the posterior.
This interpretation suggests a natural consistency requirement: if the evidence favors one token more strongly than another, it should favor the first token's complement (the event that any other token is chosen) less strongly than the second token's complement.
Prior methods use log-probability contrasts \citep{pan2026rlcsd,yu2026weak}.
Applied to our verbal feedback setup, this contrast scores each token by the change in its log-probability under the teacher when feedback is added.
However, this score can violate the consistency requirement.
Requiring this consistency, along with the other axioms of \citet{crupi2013partial}, determines the confirmation score up to a strictly increasing transformation.
Our confirmation score measures how far the feedback moves a token's probability toward zero or one, relative to the distance it had left to travel.
We then construct a target that maximizes the expected confirmation score within a trust region around the student's current distribution.
The resulting target reweights the student's own next-token probabilities with multiplicative weights that boost or discount tokens according to their scores and stay neutral when the feedback offers no guidance.

Second, OPD can still use too little of the feedback, because student rollouts limit where it can provide supervision.
Verbal feedback can provide \emph{evaluative} information about what is right or wrong in a student rollout and \emph{prescriptive} guidance on how to improve it.
Compared with feedback that only identifies an error, prescriptive guidance can more efficiently rule out competing hypotheses about what a successful rollout requires \citep{xu2026llf}.
However, OPD draws on this guidance only through the teacher's predictions at prefixes generated by the student \citep{lu2025onpolicydistillation}.
The feedback can influence these predictions while the prefix remains compatible with the suggested correction, but may become irrelevant once the student's rollout takes a different path \citep{talaei2026speculative}.
As a result, only a few token positions may carry the feedback's corrective signal, leaving much of its prescriptive content unused.
Distilling rollouts from the feedback-conditioned teacher draws on the feedback's full prescriptive content to gradually shape the student's distribution along the suggested correction.
By the time the student learns to take the corrective step, it has already received supervision on how to continue.
We bring both sources of supervision together through a simple shared-rollout estimator of a symmetric divergence between the student and target distributions over rollouts.
Using the balance heuristic \citep{veach1995optimally}, we share every student and teacher rollout across both directions, with importance ratios accounting for the pooled sampling distribution, yielding a provable bound on gradient variance.

Our contributions are as follows.
\begin{itemize}
\item \textbf{A distillation target anchored to the student.} We construct a target that isolates the change induced by verbal feedback from the teacher's pre-existing preferences. We use probabilistic confirmation \citep{crupi2013partial} to score this change and reweight the student's own distribution within a trust region. The target equals the student when feedback leaves the teacher's predictions unchanged and bounds each token's reweighting, helping stabilize training (Section~\ref{sec:iwkl}).
\item \textbf{A symmetric objective with shared rollouts.} To learn from guidance that student rollouts leave unused, we derive a shared-rollout estimator of a symmetric divergence between the student and target rollout distributions. Every student and feedback-conditioned teacher rollout contributes to both directions, providing supervision along the student's behavior and the teacher's suggested corrections. Under the student--target mixture used in our analysis, pooling prevents arbitrarily large importance weights from amplifying individual updates and gives a bound on gradient variance (Section~\ref{sec:mix}; Appendix~\ref{app:ess}).
\item \textbf{Empirical gains on knowledge and agentic benchmarks.} We evaluate our method on trivia, embodied, scientific, and tool-use benchmarks against SDPO, which follows standard feedback-conditioned OPD \citep{hubotter2026reinforcement}, and W2S-OPD, a recent contrastive variant \citep{yu2026weak}. Our method improves over both baselines (Section~\ref{sec:main-results}).
\end{itemize}

We discuss additional related work in Appendix~\ref{app:related-work}.

\section{Preliminaries}

\textbf{Notation.} A rollout $y=(y_1,\ldots,y_T)$ is a complete token sequence generated by a policy given a prompt $x$. We use $v$ to denote a candidate next token from the vocabulary. At position $t$, an autoregressive policy $\pi$ gives the next-token distribution $\pi(\cdot\mid x,y_{<t})$ at prefix $y_{<t}$. The prefix probability is $\pi(y_{<t}\mid x)=\prod_{s<t}\pi(y_s\mid x,y_{<s})$, and the complete rollout has probability $\pi(y\mid x)=\prod_{t=1}^{T}\pi(y_t\mid x,y_{<t})$.
The student policy is $\pi_\theta$, with parameters $\theta$. The operator $\mathrm{sg}(\cdot)$ preserves its input's value but blocks gradients through it.

For distributions $a$ and $b$ over the same sample space, $\mathrm{KL}(a\,\|\,b)=\mathbb{E}_{z\sim a}[\log(a(z)/b(z))]$ denotes their Kullback--Leibler divergence. We use this notation for both next-token distributions and distributions over rollouts.

\textbf{Policy gradients and token-level credit.} Given a scalar reward $R(x,y)$ for each rollout, the expected-reward objective is $J(\theta)=\mathbb{E}_{y\sim\pi_\theta(\cdot\mid x)}[R(x,y)]$.
Policy-gradient methods optimize this objective using
\begin{equation}
\label{eq:pg}
\nabla_\theta J(\theta)
=\mathbb{E}_{y\sim\pi_\theta(\cdot\mid x)}\!\left[
\sum_t A_t\,\nabla_\theta\log\pi_\theta(y_t\mid x,y_{<t})\right],
\end{equation}
where $A_t$ estimates the return following token $y_t$ relative to a baseline that does not depend on that token. The advantage assigns credit by comparing the chosen token with the alternatives at that prefix. With one reward per rollout, methods such as GRPO \citep{shao2024deepseekmath} use one group-normalized reward as the advantage of every token in a rollout. Such an update identifies which rollouts performed better, but does not distinguish the token choices responsible for that difference.

\textbf{On-policy distillation.} A teacher's next-token predictions provide one source of token-level supervision. On-policy distillation (OPD) samples rollouts from the student and, at each prefix visited, pulls the student's next-token distribution toward that of a teacher policy $\pi_T$ over the same vocabulary \citep{agarwal2024gkd, lu2025onpolicydistillation}:
\begin{equation}
\label{eq:opd}
\mathcal{L}_{\mathrm{OPD}}(\theta)
=\mathbb{E}_{y\sim\mathrm{sg}[\pi_\theta(\cdot\mid x)]}\!\left[
\sum_t\mathrm{KL}\!\left(
\pi_\theta(\cdot\mid x,y_{<t})\,\big\|\,\pi_T(\cdot\mid x,y_{<t})
\right)\right].
\end{equation}
This ordering of the KL arguments is called reverse KL, and the opposite ordering is called forward KL. At a fixed prefix, minimizing reverse KL corresponds to a policy-gradient update with token-level advantage $\log\pi_T(y_t\mid x,y_{<t})-\log\pi_\theta(y_t\mid x,y_{<t})$ in \eqref{eq:pg}; its magnitude has no uniform bound.

\textbf{Distillation from verbal feedback.} A teacher can make more informed predictions using context that the student lacks at inference, such as a reference solution, execution feedback, or verbal feedback. Conditioning the student's own model on such information can therefore make it a useful teacher \citep{hubotter2026reinforcement, zhao2026opsd, penaloza2026pidistill}. In our setting, the student generates a rollout $y^s\sim\pi_\theta(\cdot\mid x)$, and an LLM judge $\mathcal{J}$ assesses it against the task's criteria and returns verbal feedback $F=\mathcal{J}(x,y^s)$. The prompt, student rollout, and feedback form a training example $(x,y^s,F)$. The teacher conditions on this example and the current prefix to produce a next-token distribution that serves as the student's distillation target. The judge therefore need not expose token probabilities or share the student's vocabulary.
The student's predictions use only the prompt and prefix.

\section{Feedback as Evidence: A Corrected Distillation Target}
\label{sec:iwkl}

Standard feedback-conditioned OPD uses the teacher's next-token distribution directly as its distillation target. Matching that distribution can turn teacher-student differences in phrasing or formatting into training signals even when they are unrelated to the feedback or task \citep{pan2026rlcsd,yu2026weak}. We compare the teacher's predictions with and without feedback, holding the prompt $x$, student rollout $y^s$, and prefix $y_{<t}$ fixed. For each candidate token $v$, define:
\begin{equation}
\mathbb{P}_{\mathrm{post}}(v)=\pi_T(v\mid x,y^s,F,y_{<t}),
\qquad
\mathbb{P}_{\mathrm{prior}}(v)=\pi_T(v\mid x,y^s,y_{<t}).
\end{equation}
This comparison isolates the change attributable to the feedback. To turn this change into a supervision signal, we treat the feedback as evidence about the hypothesis that $v$ is the next token. The teacher supplies the prior probability $\mathbb{P}_{\mathrm{prior}}(v)$ and posterior probability $\mathbb{P}_{\mathrm{post}}(v)$ of that hypothesis. A confirmation score $S(v)$ quantifies how strongly the feedback supports or opposes that hypothesis.

\textbf{Choosing the confirmation score.} One candidate for the confirmation score is the log-probability contrast $\log\mathbb{P}_{\mathrm{post}}(v)-\log\mathbb{P}_{\mathrm{prior}}(v)$, used in prior methods \citep{ho2022classifier,pan2026rlcsd,yu2026weak}. To select our score, we use the partial-entailment framework of \citet{crupi2013partial}, which characterizes the scores satisfying four conditions, stated here for next-token hypotheses (the general probability-model form is in Appendix~\ref{app:score}):
\begin{itemize}
\item[(A0)] \emph{Probability dependence.} The score is a function of the token's probabilities with and without feedback: $S(v)=\sigma(\mathbb{P}_{\mathrm{post}}(v),\mathbb{P}_{\mathrm{prior}}(v))$.
\item[(A1)] \emph{Monotonicity.} At a fixed prior $\mathbb{P}_{\mathrm{prior}}(v)$, the score strictly increases with $\mathbb{P}_{\mathrm{post}}(v)$.
\item[(A2)] \emph{Ratio dependence under disconfirmation.} If $\mathbb{P}_{\mathrm{post}}(v)\le\mathbb{P}_{\mathrm{prior}}(v)$, the score depends only on the fraction of the prior probability that remains, $\mathbb{P}_{\mathrm{post}}(v)/\mathbb{P}_{\mathrm{prior}}(v)$.
\item[(A3)] \emph{Complementarity.} For any tokens $v$ and $v'$, the ordering of their scores reverses for their complements: $S(v)\ge S(v')$ if and only if $S(\bar v)\le S(\bar v')$, where $\bar v$ is the hypothesis that the next token is anything other than $v$. Complement scores use the same rule, $S(\bar v)=\sigma(1-\mathbb{P}_{\mathrm{post}}(v),1-\mathbb{P}_{\mathrm{prior}}(v))$.
\end{itemize}
A0 and A1 hold for both the probability difference and the log-probability difference. A2 retains the log-ratio's dependence on the probability ratio when feedback lowers a token's probability. A3 requires the confirmation scores of tokens and their complements to rank in reverse order: if one token has a higher score than another, its complement must have a lower score than the other token's complement.

\begin{theorem}[\citealp{crupi2013partial}]
\label{thm:score}
A scoring function satisfies A0--A3 for all probability models if and only if it is a strictly increasing transformation of the relative distance. Applied to the teacher's predictions, $S(v)=\psi(d(v))$ with $\psi$ strictly increasing and, for $0<\mathbb{P}_{\mathrm{prior}}(v)<1$,
\begin{equation}
\label{eq:score}
d(v)=\begin{cases}
\dfrac{\mathbb{P}_{\mathrm{post}}(v)-\mathbb{P}_{\mathrm{prior}}(v)}{1-\mathbb{P}_{\mathrm{prior}}(v)} & \text{if } \mathbb{P}_{\mathrm{post}}(v)\ge\mathbb{P}_{\mathrm{prior}}(v),\\[6pt]
\dfrac{\mathbb{P}_{\mathrm{post}}(v)-\mathbb{P}_{\mathrm{prior}}(v)}{\mathbb{P}_{\mathrm{prior}}(v)} & \text{otherwise}.
\end{cases}
\end{equation}
\end{theorem}
The relative distance measures how far the feedback moves a token's probability toward zero or one, relative to the distance it had left to travel. For an increase, the probability gained is divided by $1-\mathbb{P}_{\mathrm{prior}}(v)$; for a decrease, the signed change is divided by $\mathbb{P}_{\mathrm{prior}}(v)$. The theorem fixes the ranking and leaves the numerical scale free. We take $\psi$ to be the identity, giving a signed score $S=d\in[-1,1]$ that is zero when the feedback leaves the token's probability unchanged.

\begin{figure}[t]
    \centering
    \includegraphics[width=\linewidth]{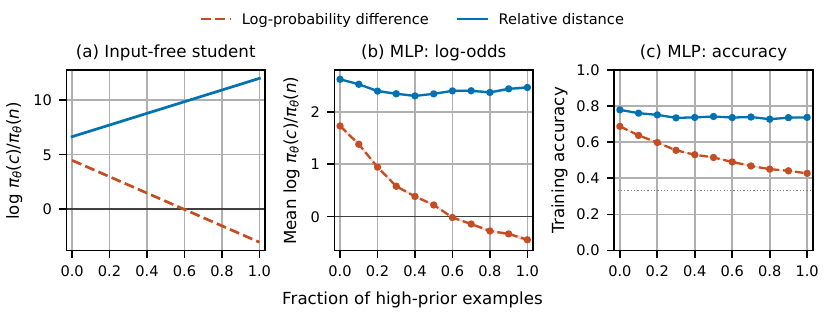}
    \caption{\textbf{The log-probability difference can make the student prefer a wrong token even when feedback reinforces the correct one.} In a synthetic three-token task, we compare two scores for constructing training targets by reweighting the student's probabilities. Both variants minimize reverse KL to their respective targets and differ only in the confirmation score. We vary the fraction of high-prior examples (horizontal axis), on which feedback raises the teacher's probability of the correct token $c$ from $95\%$ to $98\%$ and of a wrong token $n$ from $1\%$ to $1.2\%$. The log-probability difference scores $n$ higher because $n$'s proportional increase is larger. Our relative-distance score measures each increase as a fraction of the probability left to gain, giving $c$ the higher score. \textbf{(a)}~Student log-odds of $c$ over $n$ under exact optimization of a single distribution shared across examples. \textbf{(b)}~Mean log-odds across training examples for an MLP student. Negative log-odds indicate a preference for the wrong token. \textbf{(c)}~Fraction of training examples on which the MLP assigns the highest probability to $c$ among the three tokens; the dotted line marks random-choice accuracy ($1/3$). Panels (b,c) show means $\pm$ standard errors over five paired seeds. Experimental details and the connection to A3 are in Appendix~\ref{app:score-reversal}.}
    \label{fig:score-reversal}
\end{figure}

The log-probability difference satisfies A0--A2 but violates A3. For example, it ranks a token whose probability increases from $0.01$ to $0.02$ above one whose probability increases from $0.40$ to $0.60$. Scoring their complements gives the same order, whereas A3 requires the reverse. The relative-distance score gives reverse rankings as required (Appendix~\ref{app:score}). A strictly increasing transformation preserves the log-ratio's rankings, so rescaling cannot repair this failure.
Figure~\ref{fig:score-reversal} shows that this violation can reverse the student's preference for the correct token, even under exact optimization.

\textbf{Constructing the target.} 
We now use the confirmation score to specify how the student's predictions should change. At each prefix, we use the current student's next-token distribution as a reference. For a reweighting strength $\beta>0$, we choose the corrected target $\pi^{\mathrm{corr}}$ to maximize the expected score while penalizing KL divergence from this reference. This gives a KL-regularized trust-region problem over candidate next-token distributions $\pi$:
\begin{equation}
\label{eq:target-variational}
\pi^{\mathrm{corr}}(\cdot\mid x,y^s,F,y_{<t})
=\operatorname*{arg\,max}_{\pi}
\left\{\mathbb{E}_{v\sim\pi}[S(v)]
-\frac{1}{\beta}\mathrm{KL}\!\left(\pi\,\big\|\,\pi_\theta(\cdot\mid x,y_{<t})\right)\right\}.
\end{equation}
This improvement step \citep{rafailov2024direct} has a closed-form solution that reweights the student's own next-token probabilities, with normalizing constant $Z_t$:
\begin{equation}
\label{eq:iwkl}
\begin{aligned}
\pi^{\mathrm{corr}}(v\mid x,y^s,F,y_{<t})
&=\mathrm{sg}\!\left[\frac{\pi_\theta(v\mid x,y_{<t})\exp\{\beta S(v)\}}{Z_t}\right],\\
Z_t&=\sum_v\pi_\theta(v\mid x,y_{<t})\exp\{\beta S(v)\}.
\end{aligned}
\end{equation}
Before normalization, positive scores upweight tokens, negative scores downweight them, and a zero score leaves the student's probability for that token unchanged. A token that the feedback fully confirms or refutes receives a weight of $e^{\beta}$ or $e^{-\beta}$, respectively. The target therefore starts from the student's own predictions and adjusts them according to how the feedback changes the teacher's predictions.

\textbf{Properties of the target.} When the feedback leaves the teacher's entire next-token distribution unchanged at a prefix, $\mathbb{P}_{\mathrm{post}}=\mathbb{P}_{\mathrm{prior}}$ and $S=0$. The target then equals the current student's next-token distribution, so the prefix contributes no local distillation gradient. Moreover, because $S\in[-1,1]$,
\begin{equation}
\label{eq:target-ratio-bound}
e^{-2\beta}\le\frac{\pi^{\mathrm{corr}}(v\mid x,y^s,F,y_{<t})}{\pi_\theta(v\mid x,y_{<t})}\le e^{2\beta}
\end{equation}
for every token in the student's support (Appendix~\ref{app:target}). Each target probability therefore remains within a fixed multiplicative factor of the current student's probability, providing control over the strength of the distillation signal.

The corrected target can be used with different distillation objectives. For example, under reverse KL OPD, it yields a policy-gradient update with bounded token-level advantage $\beta\,(S(v)-\mathbb{E}_{v'\sim\pi_\theta(\cdot\mid x,y_{<t})}[S(v')])$ (Appendix~\ref{app:target}). In the next section, we combine this corrected target with a shared objective to also learn from guidance that student rollouts leave unused.

\section{A Shared Objective for Student and Teacher Rollouts}
\label{sec:mix}

Standard OPD concentrates supervision on prefixes the student actually visits. A student rollout can leave most of the feedback unused even when the teacher understands the suggested correction. While the student's prefix remains compatible with that correction, feedback can shift the teacher's predictions toward continuations that carry it out. Once the student rollout takes a different path, the feedback may become less relevant to the prefixes that follow \citep{talaei2026speculative}. Its prescriptive content may therefore affect only a few token positions, consistent with findings that useful supervision in OPD can be concentrated at relatively few token positions \citep{xie2026position,xu2026tip}. We add feedback-conditioned teacher rollouts to provide supervision along the suggested corrections. We derive a shared-rollout estimator of a symmetric divergence over rollouts using the balance heuristic \citep{veach1995optimally}, with every student and teacher rollout contributing to both directions and importance ratios accounting for the pooled sampling distribution.

\textbf{A rollout-level objective.} Fix a training example $(x,y^s,F)$. The corrected next-token targets define a rollout distribution $\pi^{\mathrm{corr}}(\cdot\mid x,y^s,F)$.
We want to retain OPD supervision at prefixes visited by the student while also supervising it at prefixes generated by the corrected policy.
Combining the reverse and forward KL terms provides both sources of supervision, giving the Jeffreys divergence:
\begin{equation}
\label{eq:jeffreys-seq}
\begin{aligned}
\mathrm{J}(\pi_\theta,\pi^{\mathrm{corr}})
&=\mathrm{KL}\!\left(\pi_\theta(\cdot\mid x)\,\big\|\,\pi^{\mathrm{corr}}(\cdot\mid x,y^s,F)\right)\\
&\quad+\mathrm{KL}\!\left(\pi^{\mathrm{corr}}(\cdot\mid x,y^s,F)\,\big\|\,\pi_\theta(\cdot\mid x)\right),
\end{aligned}
\end{equation}
averaged over training examples.
By the KL chain rule,
\begin{equation}
\label{eq:jeffreys-tok}
\begin{aligned}
\mathrm{J}(\pi_\theta,\pi^{\mathrm{corr}})
&=\mathbb{E}_{y\sim\pi_\theta(\cdot\mid x)}\!\left[\sum_t
\mathrm{KL}\!\left(\pi_\theta(\cdot\mid x,y_{<t})\,\big\|\,\pi^{\mathrm{corr}}(\cdot\mid x,y^s,F,y_{<t})\right)\right]\\
&\quad+\mathbb{E}_{y\sim\pi^{\mathrm{corr}}(\cdot\mid x,y^s,F)}\!\left[\sum_t
\mathrm{KL}\!\left(\pi^{\mathrm{corr}}(\cdot\mid x,y^s,F,y_{<t})\,\big\|\,\pi_\theta(\cdot\mid x,y_{<t})\right)\right].
\end{aligned}
\end{equation}
The reverse term is the OPD loss~\eqref{eq:opd} with $\pi^{\mathrm{corr}}$ as the target, providing supervision at prefixes the student currently visits. The forward term provides supervision at prefixes generated by $\pi^{\mathrm{corr}}$, shaping the student's predictions even along rollouts it cannot yet reliably generate.

\textbf{One pool for both directions.} With separate pools, student rollouts estimate only the reverse KL term, and rollouts from the corrected policy estimate only the forward KL term. We share every rollout across both terms, with importance weights recovering the prefix distribution required by each term \citep{veach1995optimally,hesterberg1995defensive}. We generate each rollout entirely from $\pi_\theta$ with probability $\lambda$ or from $\pi^{\mathrm{corr}}$ with probability $1-\lambda$, for a fixed $0<\lambda<1$. The source is chosen once per rollout, giving the rollout distribution $m(\cdot\mid x,y^s,F)=\lambda\pi_\theta(\cdot\mid x)+(1-\lambda)\pi^{\mathrm{corr}}(\cdot\mid x,y^s,F)$. Its prefix probabilities are
\[
m(y_{<t}\mid x,y^s,F)=\lambda\pi_\theta(y_{<t}\mid x)+(1-\lambda)\pi^{\mathrm{corr}}(y_{<t}\mid x,y^s,F).
\]
Sampling from this mixture requires no custom decoder to combine the rollout sources (Appendix~\ref{sec:impl}).
We recover each term of \eqref{eq:jeffreys-tok} by weighting prefixes from this mixture with the appropriate importance ratio:
\begin{equation}
\label{eq:balance}
w_{\mathrm{rev}}(y_{<t})=\frac{\pi_\theta(y_{<t}\mid x)}{m(y_{<t}\mid x,y^s,F)}\le\frac{1}{\lambda},
\qquad
w_{\mathrm{fwd}}(y_{<t})=\frac{\pi^{\mathrm{corr}}(y_{<t}\mid x,y^s,F)}{m(y_{<t}\mid x,y^s,F)}\le\frac{1}{1-\lambda}.
\end{equation}
Because the mixture contains both policies, the importance ratios are bounded by construction. These bounds avoid the arbitrarily large ratios that can arise from a direct importance ratio between the two policies. Applying the change of sampling distribution to each term of \eqref{eq:jeffreys-tok} gives the shared-rollout form
\begin{equation}
\label{eq:pooled}
\begin{aligned}
\mathrm{J}(\pi_\theta,\pi^{\mathrm{corr}})
=\mathbb{E}_{y\sim m(\cdot\mid x,y^s,F)}\sum_t\Big[&
w_{\mathrm{rev}}(y_{<t})\,\mathrm{KL}\!\left(\pi_\theta(\cdot\mid x,y_{<t})\,\big\|\,\pi^{\mathrm{corr}}(\cdot\mid x,y^s,F,y_{<t})\right)\\
{}+{}&w_{\mathrm{fwd}}(y_{<t})\,\mathrm{KL}\!\left(\pi^{\mathrm{corr}}(\cdot\mid x,y^s,F,y_{<t})\,\big\|\,\pi_\theta(\cdot\mid x,y_{<t})\right)\Big].
\end{aligned}
\end{equation}
As in OPD, training uses a semi-gradient, holding the target, sampled prefixes, and importance ratios fixed while differentiating the token losses.

At each position, sharing rollouts gives each term a strictly larger asymptotic effective sample size than separate pools with the same rollout allocation, unless the student and target prefix distributions have disjoint support. The bounded importance ratios also give a variance bound for the estimated semi-gradient with respect to the student's parameters. Appendix~\ref{app:ess} establishes both results.

For efficient rollout generation, we use the feedback-conditioned teacher as a proposal for the corrected target. Appendix~\ref{sec:impl} describes the resulting importance weights and implementation details.

\section{Results}

Having isolated the benefit of the confirmation score under reverse KL alone (Figure~\ref{fig:score-reversal}), we now evaluate the corrected target together with our shared-rollout objective. These experiments assess how the complete method learns from verbal feedback across knowledge and agentic benchmarks.

\subsection{Experimental Setup}
\label{sec:experimental-setup}

\textbf{Benchmarks.} Trivia Fantasy tests whether the student can acquire 20 fictional facts initially unknown to both the student and teacher, with all new factual information supplied through the judge's feedback. On some questions, the student initially assigns near-zero probability to the correct answer. We measure acquisition through recall on the training questions without feedback. 

We also evaluate on three benchmarks with held-out tasks. EmbodiedEval is our internal benchmark of an embodied planner's compliance with behavioral criteria, none of which have programmatic verifiers. ToolAlpaca~\citep{tang2023toolalpaca} tests tool use on APIs absent from training, and SciKnowEval~\citep{feng2024sciknoweval} tests Level-3 scientific reasoning across four domains. Dataset construction, splits, and scoring are detailed in Appendix~\ref{app:benchmarks}.

\textbf{Training.} We compare methods under a common supervision interface in which an LLM judge provides verbal feedback during training. Within each benchmark, all methods share the student initialization, feedback judge, training data, and optimizer settings. We train each method for 200 steps or until the evaluation score plateaus. Unless stated otherwise, the ablations use the same setup. Model choices, hyperparameters, and implementation details are given in Appendix~\ref{app:exp-details}.

\textbf{Evaluation and reporting.} On ToolAlpaca and SciKnowEval, we follow SDPO's avg@16: each held-out question is answered 16 times, and the reported score averages the per-question accuracies. Each entry in Table~\ref{tab:main} is the mean of three checkpoint evaluations, with a 95\% confidence interval from the $t$-distribution with two degrees of freedom. We evaluate the base model with frozen weights and report the mean with a $t$-based 95\% confidence interval over independent passes. Sampling settings, evaluation intervals, and the number of base-model evaluation passes are listed in Table~\ref{tab:hparams-bench}.

\textbf{Baselines.} Self-Distillation Policy Optimization (SDPO)~\citep{hubotter2026reinforcement} learns from privileged information by matching a feedback-conditioned self-teacher on student-generated rollouts. We adapt it to our judge's verbal feedback and use the frozen-initial-teacher variant considered in the original work. Weak-to-Strong On-Policy Distillation (W2S-OPD)~\citep{yu2026weak} constructs a proxy teacher by applying a contrastive log-probability shift to a frozen copy of the initial student. We adapt its contrast pair to compare the same teacher with and without the judge's verbal feedback along student-generated rollouts.

\subsection{Main Results}
\label{sec:main-results}

\begin{table}[htbp]
\centering
\normalsize
\setlength{\tabcolsep}{2.5pt}
\renewcommand{\arraystretch}{1.7}
\newcommand{\mainresult}[2]{$#1$\,{\scriptsize\color{black!65}$\pm #2$}}
\begin{tabularx}{\textwidth}{lcc*{4}{>{\centering\arraybackslash}X}c}
\toprule
\multirow{2}{*}{\textbf{Method}}
& \multirow{2}{*}{\shortstack{\textbf{Trivia}\\\textbf{Fantasy}}}
& \multirow{2}{*}{\textbf{EmbodiedEval}}
& \multicolumn{4}{c}{\textbf{SciKnowEval (L3)}}
& \multirow{2}{*}{\textbf{ToolAlpaca}} \\
\cmidrule(lr){4-7}
& & & Chem. & Phys. & Biology & Materials & \\
\midrule
Base model & \mainresult{\phantom{0}2.8}{1.9} & \mainresult{57.3}{2.0} & \mainresult{31.9}{0.3} & \mainresult{53.8}{0.1} & \mainresult{22.1}{4.1} & \mainresult{68.4}{0.8} & \mainresult{47.6}{1.7} \\
\midrule
SDPO       & \mainresult{61.0}{7.1} & \mainresult{73.7}{1.0} & \mainresult{35.9}{3.4} & \mainresult{54.9}{1.9} & \mainresult{27.5}{0.7} & \mainresult{73.2}{0.3} & \mainresult{59.5}{0.5} \\
W2S-OPD    & \mainresult{41.7}{7.2} & \mainresult{60.8}{4.2} & \mainresult{35.0}{1.1} & \mainresult{54.8}{0.4} & \mainresult{27.3}{1.9} & \mainresult{72.1}{0.4} & \mainresult{59.7}{2.0} \\
\addlinespace[2pt]
\rowcolor{blue!4}
\textbf{Ours} & \mainresult{\mathbf{98.3}}{2.9} & \mainresult{\mathbf{82.1}}{3.0} & \mainresult{\mathbf{50.2}}{2.9} & \mainresult{\mathbf{66.9}}{1.6} & \mainresult{\mathbf{28.2}}{1.8} & \mainresult{\mathbf{73.7}}{1.5} & \mainresult{\mathbf{61.9}}{2.4} \\
\bottomrule
\end{tabularx}
\caption{\textbf{Our method achieves the highest mean score across all benchmarks and SciKnowEval subjects.} Scores (\%; higher is better) are means with 95\% confidence intervals; the highest mean in each column is bold. Trained-model scores are averaged over three checkpoint evaluations; base-model scores are averaged over independent evaluation passes (Section~\ref{sec:experimental-setup}).}
\label{tab:main}
\end{table}

Our method achieves the highest mean score across all four benchmarks, including each SciKnowEval subject (Table~\ref{tab:main}). The largest gain is on Trivia Fantasy, where accuracy rises from the base model's $2.8\%$ to $98.3\%$, compared with $61.0\%$ for SDPO and $41.7\%$ for W2S-OPD. This near-perfect accuracy demonstrates that the student can learn the judge's answer key through verbal feedback interpreted by a teacher that initially lacks those facts.

The gains also extend to held-out tasks. On EmbodiedEval, the mean pass rate rises from SDPO's $73.7\%$ to $82.1\%$, reducing the mean task failure rate by roughly a third. On SciKnowEval, our method exceeds the stronger baseline by $14.3$ percentage points in chemistry and $12.0$ in physics. Physics is particularly informative: both baselines stay within $1.1$ points of the base model's $53.8\%$, whereas our method reaches $66.9\%$.

The gains are smaller in biology, materials science, and tool use: $0.7$, $0.5$, and $2.2$ percentage points over the stronger baseline, respectively. The reported confidence intervals overlap in these comparisons, so the methods are less clearly separated on these tasks.

\subsection{Ablations}
\label{sec:ablations}
\label{sec:exp-pool}
\label{sec:exp-beta}

We study rollout allocation and target reweighting in 100-step ToolAlpaca training runs, using the setup in Section~\ref{sec:experimental-setup} and seeded evaluation sampling. ToolAlpaca measures performance on the trained task, while the four SciKnowEval Level-3 subjects measure zero-shot transfer.

\textbf{Rollout allocation.} We vary the student share $\lambda$ of the rollout pool in Section~\ref{sec:mix} over $\lambda\in\{1/128,\,1/4,\,1/2,\,3/4,\,127/128\}$, keeping the total rollout budget fixed.

\begin{figure}[t]
\centering
\includegraphics[width=\linewidth]{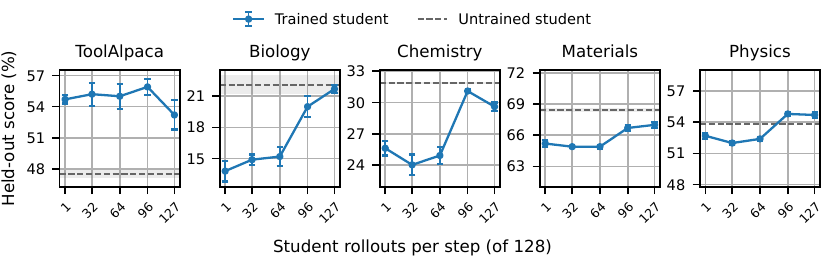}
\caption{\textbf{A pool of 96 student and 32 teacher rollouts gives the highest mean ToolAlpaca score while keeping transfer scores near the untrained student.} The x-axis shows the number of student rollouts per step out of a fixed total of 128; the remaining rollouts come from the teacher. We train on ToolAlpaca for 100 steps. The left panel evaluates held-out ToolAlpaca tasks; the others evaluate zero-shot transfer to SciKnowEval. Scores average accuracy over 16 sampled rollouts per question. Points show means over three checkpoints; error bars show standard errors. Dashed lines and shaded bands show the untrained student (mean $\pm$ standard error over three independent evaluation passes).}
\label{fig:pool-lambda}
\end{figure}

Every tested mixture improves ToolAlpaca, but similar task scores can conceal substantially different transfer costs (Figure~\ref{fig:pool-lambda}). ToolAlpaca varies by only $2.7$ percentage points across the sweep, while biology and chemistry vary by $7.9$ and $7.1$ points, respectively.

With a quarter of the pool drawn from the teacher ($\lambda=3/4$), ToolAlpaca reaches its highest mean score, $55.9\%$, and every SciKnowEval subject stays within $2.1$ points of the untrained student. Increasing the teacher share beyond this point provides no further gain on ToolAlpaca and lowers performance on all four transfer subjects. At the other extreme, using just one teacher rollout per 128 still improves ToolAlpaca by $5.7$ points over the untrained student.

\textbf{Reweighting strength.} We vary the reweighting strength $\beta$ of the corrected target in Section~\ref{sec:iwkl} over $\beta\in\{1,\,2,\,5,\,10\}$.

\begin{figure}[t]
\centering
\includegraphics[width=\linewidth]{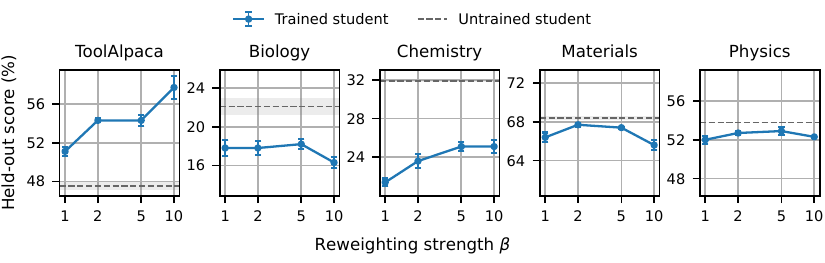}
\caption{\textbf{Stronger feedback reweighting improves ToolAlpaca, with mixed effects on transfer.} We vary the reweighting strength $\beta$ of the corrected target and train on ToolAlpaca for 100 steps. The left panel evaluates held-out ToolAlpaca tasks; the others evaluate zero-shot transfer to SciKnowEval. Scores average accuracy over 16 sampled rollouts per question. Points and error bars show the mean $\pm$ standard error over three checkpoint evaluations. Dashed lines and shaded bands show the untrained student (mean $\pm$ standard error over three independent evaluation passes).}
\label{fig:beta}
\end{figure}

At the same 100-step training budget, increasing $\beta$ from $1$ to $10$ raises ToolAlpaca accuracy from $51.1\%$ to $57.7\%$, a gain of $6.6$ percentage points (Figure~\ref{fig:beta}). The improvement is uneven: $\beta=2$ and $\beta=5$ both reach $54.3\%$, followed by a further gain at $\beta=10$.

The effect of reweighting strength on transfer varies by subject. Chemistry improves by $3.7$ points from $\beta=1$ to $10$, and physics changes little, whereas biology and materials decline at the upper end of the sweep. Transfer scores remain below the untrained student across the sweep.

%

\FloatBarrier
\subsection*{Acknowledgments}
This work is supported in part by the US National Science Foundation under the awards III-2128019, SLES-2331904, and CAREER-2442098, the Commonwealth Cyber Initiative's Central Virginia Node under awards VV-1Q26-001, VV-1Q27-007, and VV-1Q27005, 4-VA at UVA, a Cisco Faculty Research Award, and a Google Research Award. We thank Craig Friedman from AWS for providing support in obtaining compute resources for the project.

\bibliography{bibliography}
\bibliographystyle{iclr2027_conference}

\clearpage

\appendix
\setcounter{figure}{0}
\renewcommand{\thefigure}{A.\arabic{figure}}
\renewcommand{\theHfigure}{appendix.\arabic{figure}}
\section{Related Work}
\label{app:related-work}
On-policy distillation has emerged as an efficient approach to language-model post-training, matching a teacher's predictions along the student's own rollouts \citep{yang2025qwen3,lu2025onpolicydistillation}.
Instead of distilling a separate teacher, the student can supervise itself when given privileged information such as demonstrations or reference solutions \citep{zhao2026opsd,shenfeld2026self}.
Conditioning the teacher on verbal feedback extends this approach to learning from critiques of the student's behavior \citep{hubotter2026reinforcement,song2026rltf}.
Our work separates the feedback's guidance from the teacher's existing preferences and learns from corrections that student rollouts may fail to follow.

To separate useful guidance from teacher preferences, prior work removes signals shared across questions \citep{shen2026generic} or compares predictions conditioned on correct and incorrect solutions \citep{pan2026rlcsd,ogls2026}.
Concurrent work uses teachers to encourage or discourage behaviors \citep{baumann2026repulsive} and adjusts predictions for differences between teacher and student contexts \citep{he2026calibrating}.
Our comparison holds the question and response fixed and changes only the feedback.
As with prior targets that reweight a fixed model's predictions using log-probability differences \citep{yu2026weak}, it requires no correctness labels.
We instead reweight the student's current predictions using a bounded confirmation score that ranks tokens and their complements in reverse order \citep{crupi2013partial}.
Other work matches the teacher's predictions only where privileged information increases the generated token's probability \citep{lazaridis2026edge}; we use both increases and decreases to construct the target.
The target equals the student when feedback leaves the teacher's predictions unchanged.

Combining student and teacher rollouts provides supervision at prefixes the student rarely visits \citep{lazaridis2026edge}.
Prior work chooses the KL direction according to which model generated each response or dialogue turn \citep{ko2025distillm2,li2026guided}.
Our estimator instead shares every student and target rollout across both KL directions, with separate importance ratios recovering the prefix distribution required by each direction.

\section{Proofs and derivations}
\label{app:proofs}
\textbf{Appendix notation.} For the rollout analysis, let $\pi_{\mathrm{tar}}$ denote a target policy, such as the teacher distribution, the W2S-OPD target \citep{yu2026weak}, or our corrected target $\pi^{\mathrm{corr}}$. Throughout the appendix, we abbreviate the student and target policies as $q=\pi_\theta$ and $p=\pi_{\mathrm{tar}}$ and suppress the fixed conditioning context $(x,y^s,F)$. At a prefix $u=y_{<t}$, write $q_t=\pi_\theta(\cdot\mid x,u)$ and $p_t=\pi_{\mathrm{tar}}(\cdot\mid x,y^s,F,u)$ for the next-token distributions, and $q(u)=\pi_\theta(u\mid x)$, $p(u)=\pi_{\mathrm{tar}}(u\mid x,y^s,F)$, and $m(u)=m(u\mid x,y^s,F)$ for prefix probabilities. The notation $u\sim q$, $u\sim p$, or $u\sim m$ refers to the corresponding prefix law at a fixed position, while $q_{\mathrm{seq}}$, $p_{\mathrm{seq}}$, and $m_{\mathrm{seq}}$ denote complete-rollout distributions. We write $w_q=w_{\mathrm{rev}}$ and $w_p=w_{\mathrm{fwd}}$ for the importance ratios in \eqref{eq:balance}, and $p^{\mathrm{corr}}=\pi^{\mathrm{corr}}$ for the corrected target, using the same token, prefix, and rollout conventions. For the feedback-conditioned teacher, write $\pi_F(\cdot\mid u):=\pi_T(\cdot\mid x,y^s,F,u)$ for its next-token distribution and $\pi_F(u)$ for its prefix probability.

At each fixed prefix, we also abbreviate the teacher's prior and posterior probabilities as $b(v)=\mathbb{P}_{\mathrm{prior}}(v)$ and $f(v)=\mathbb{P}_{\mathrm{post}}(v)$, respectively. We use these shorthands throughout the appendix.

\subsection{Confirmation axioms and the relative-distance score}
\label{app:score}
We state the abstract probability conditions used to select the score; applying that score to teacher predictions requires only $f(v)$ and $b(v)$. Let $H,E$ be nontrivial events in a finite event space, and let $P$ be a probability law that assigns positive mass to each atom. Write $C_P(H,E)$ for the confirmation score of hypothesis $H$ given evidence $E$. The following conditions are imposed across all such event spaces, events, and laws:
\begin{itemize}
\item[(A0)] There is a common function $\sigma$ such that $C_P(H,E)=\sigma(P(H\mid E),P(H))$.
\item[(A1)] For every $b\in(0,1)$, $\sigma(a,b)$ is strictly increasing in $a\in[0,1]$.
\item[(A2)] If $P(H\mid E)\le P(H)$, then $C_P(H,E)=C_P(E,H)$.
\item[(A3)] $C_P(H_1,E)\ge C_P(H_2,E)$ if and only if $C_P(H_1^c,E)\le C_P(H_2^c,E)$.
\end{itemize}
\textbf{Connection to the cited theorem.} These conditions specialize the representation theorem of \citet{crupi2013partial}, incorporating its 2014 erratum. Their formality condition permits dependence on the triple $(P(H\cap E),P(H),P(E))$. Our A0 implies theirs by taking $g(j,h,e)=\sigma(j/e,h)$; A1 implies their ordering condition at fixed $H$ and $P$, and A2--A3 coincide with theirs.

Under our A0, A2 is equivalent to ratio dependence on the disconfirmation branch, as stated in the main text. Ratio dependence implies A2 because $P(H\mid E)/P(H)=P(E\mid H)/P(E)$. Conversely, fix $r\in[0,1]$ and any two priors $b_1,b_2\in(0,1)$. Choose $e>0$ smaller than both $(1-b_i)/(1-rb_i)$. For each $i$, the assignments $P(H)=b_i$, $P(E)=e$, and $P(H\cap E)=rb_i e$ are coherent. A2 gives $\sigma(rb_i,b_i)=\sigma(re,e)$, so the score depends only on $r$. If $r=0$, take $H$ and $E$ disjoint; all nonempty atoms still have positive mass.

The cited theorem therefore implies that $\sigma$ is a strictly increasing transformation of the relative distance in \eqref{eq:score}. Conversely, every such transformation satisfies our A0--A3, establishing Theorem~\ref{thm:score}.

\textbf{Why the log-ratio fails complementarity.} Consider three tokens with
\begin{equation}
b=(0.01,\,0.40,\,0.59),
\qquad
f=(0.02,\,0.60,\,0.38).
\end{equation}
For the first two tokens, the log-ratio scores satisfy $\log(0.02/0.01)=\log2>\log1.5=\log(0.60/0.40)$. Their complements have the same ordering: $\log(0.98/0.99)>\log(0.40/0.60)$. This violates A3. A strictly increasing transformation cannot change either ordering. These distributions also admit a common evidence event: choose $P(E)=1/4$, $P(v,E)=f(v)/4$, and $P(v,E^c)=b(v)-f(v)/4$, which are all positive. In contrast, relative distance gives scores $1/99<1/3$ for the two tokens and $-1/99>-1/3$ for their complements. More generally, $d(1-f,1-b)=-d(f,b)$.

\subsection{Properties of the corrected target}
\label{app:target}
At a fixed nonterminal prefix, let $\bar q=\mathrm{sg}(q_t)$ denote the student distribution held fixed when constructing the target, and write $p^{\mathrm{corr}}$ for the corrected next-token distribution. Assume $\bar q(v)>0$ and $0<b(v)<1$ over the vocabulary. For a fixed score $S$, the normalized target in \eqref{eq:iwkl} satisfies
\begin{equation}
\mathbb{E}_{v\sim\pi}[S(v)]-\frac{1}{\beta}\mathrm{KL}(\pi\,\|\,\bar q)
=\frac{1}{\beta}\log Z_t-\frac{1}{\beta}\mathrm{KL}(\pi\,\|\,p^{\mathrm{corr}}).
\end{equation}
Nonnegativity of KL proves that $p^{\mathrm{corr}}$ solves \eqref{eq:target-variational}. For $S=d$, the score lies in $[-1,1]$, so $e^{-\beta}\le Z_t\le e^\beta$. Substitution into \eqref{eq:iwkl} gives \eqref{eq:target-ratio-bound}. If $f(v)=b(v)$ for every token, then $S\equiv0$, $Z_t=1$, and $p^{\mathrm{corr}}=\bar q$. At $q=\bar q$, both token KL terms, including their skew versions, are minimized and have zero local gradient when the target is held fixed. At absorbing end-of-sequence prefixes, the end-of-sequence token is repeated deterministically and contributes zero loss.

\textbf{Policy-gradient form of the token KL terms.} Let $z$ be the logits of $q_t$, so that $q_t(v)=\mathrm{softmax}(z)_v$ and $\partial q_t(v)/\partial z_k=q_t(v)\,(\mathbf{1}[v=k]-q_t(k))$. For a fixed target $p$ and $\ell(v)=\log q_t(v)-\log p(v)$,
\begin{equation}
\frac{\partial}{\partial z_k}\mathrm{KL}(q_t\,\|\,p)=q_t(k)\big[\ell(k)-\mathbb{E}_{q_t}[\ell]\big],
\qquad
\frac{\partial}{\partial z_k}\mathrm{KL}(p\,\|\,q_t)=q_t(k)-p(k).
\end{equation}
With $p=p^{\mathrm{corr}}$ and $q_t=\bar q$ at the evaluation point, $\ell(v)=-\beta S(v)+\log Z_t$, so the first expression equals $-\beta\,\bar q(k)\big(S(k)-\mathbb{E}_{\bar q}[S]\big)=-\beta\,\partial\,\mathbb{E}_{q_t}[S]/\partial z_k$, treating $S$ as fixed. The semi-gradient of the reverse term is therefore $-\beta$ times the policy gradient of the expected score, with per-token advantage $\beta(S(k)-\mathbb{E}_{\bar q}[S])\in[-2\beta,2\beta]$. The second expression equals $\bar q(k)\big(1-e^{\beta S(k)}/Z_t\big)$, so descending the forward term applies the advantage $e^{\beta S(k)}/Z_t-1$, which increases with $S$, has zero mean under $\bar q$, and is bounded by \eqref{eq:target-ratio-bound}. The chain rule through $z=z(\theta)$ carries both statements to parameter gradients.

\subsection{Pooled estimator and gradient variance}
\label{app:ess}
Fix the conditioning context $(x,y^s,F)$, a finite vocabulary, and a common finite horizon $T$ with absorbing end-of-sequence padding. Sample fresh rollouts from $m_{\mathrm{seq}}=\lambda q_{\mathrm{seq}}+(1-\lambda)p_{\mathrm{seq}}$. At each position, $q$, $p$, and $m=\lambda q+(1-\lambda)p$ denote the corresponding prefix laws, with importance ratios defined where $m(u)>0$. The KL and gradient results assume common rollout support and finite token gradients; the change-of-measure and ESS identities allow arbitrary support.

For autoregressive rollout distributions, $\log(q_{\mathrm{seq}}(y)/p_{\mathrm{seq}}(y))=\sum_t\log(q_t(y_t)/p_t(y_t))$. Taking expectation under $q_{\mathrm{seq}}$ and conditioning on each prefix gives
\begin{equation}
\mathrm{KL}(q_{\mathrm{seq}}\,\|\,p_{\mathrm{seq}})
=\sum_t\mathbb{E}_{u\sim q}\mathbb{E}_{v\sim q_t}
\left[\log\frac{q_t(v)}{p_t(v)}\right]
=\sum_t\mathbb{E}_{u\sim q}\big[\mathrm{KL}(q_t\,\|\,p_t)\big].
\end{equation}
Interchanging $q$ and $p$ proves \eqref{eq:jeffreys-tok}. For any integrable prefix function $h$ and component $\ell\in\{q,p\}$, $\mathbb{E}_{u\sim m}[w_\ell(u)h(u)]=\sum_u\ell(u)h(u)=\mathbb{E}_{u\sim\ell}[h(u)]$. Applying this change of measure to each token divergence proves \eqref{eq:pooled}.

\textbf{Second moment of the combined gradient.} Fix a position $t$ and let $g_q(u)=\nabla_\theta\mathrm{KL}(q_t\,\|\,p_t)$ and $g_p(u)=\nabla_\theta\mathrm{KL}(p_t\,\|\,q_t)$ be the unweighted token gradients at prefix $u$, with the target held fixed. Their second moments are finite under the preceding assumptions; use the Euclidean norm. Holding the importance weights fixed, the combined contribution is $G(u)=w_q(u)g_q(u)+w_p(u)g_p(u)$. Define $r(u)=\lambda w_q(u)$; then $1-r(u)=(1-\lambda)w_p(u)$ and
\begin{equation}
\label{eq:pooled-gradient-convex}
G(u)=r(u)\frac{g_q(u)}{\lambda}
+(1-r(u))\frac{g_p(u)}{1-\lambda}.
\end{equation}
For $\lambda=1/2$, $G/2$ is a convex combination of the two component gradients, so $\|G\|\le2\max\{\|g_q\|,\|g_p\|\}$. More generally, convexity of the squared norm in \eqref{eq:pooled-gradient-convex}, followed by $m r=\lambda q$ and $m(1-r)=(1-\lambda)p$, gives
\begin{equation}
\label{eq:pooled-gradient-moment}
\mathbb{E}_m\|G\|^2
\le B:=\frac{\mathbb{E}_q\|g_q\|^2}{\lambda}
+\frac{\mathbb{E}_p\|g_p\|^2}{1-\lambda}.
\end{equation}
The multiplicative factors depend only on the mixture proportions. The same argument applies to the skew token gradients in Appendix~\ref{sec:impl}.

\textbf{Variance of rollout updates.} Restore the position index and write $B_t$ for the bound in \eqref{eq:pooled-gradient-moment}. A rollout contributes $H(Y)=\sum_{t=1}^T G_t(Y_{<t})$. Minkowski's inequality gives $\mathbb{E}_{m_{\mathrm{seq}}}\|H\|^2\le(\sum_t\sqrt{B_t})^2$. For $N$ independent rollouts from $m_{\mathrm{seq}}$, let $\widehat H_N=N^{-1}\sum_{i=1}^N H(Y_i)$ and $\mu=\mathbb{E}_{m_{\mathrm{seq}}}H$. The sum of the coordinate variances of their average satisfies
\begin{equation}
\label{eq:pooled-gradient-variance}
\mathbb{E}\|\widehat H_N-\mu\|^2
=\frac{\mathbb{E}_{m_{\mathrm{seq}}}\|H\|^2-\|\mu\|^2}{N}
\le\frac{\big(\sum_t\sqrt{B_t}\big)^2-\|\mu\|^2}{N}
\le\frac{T}{N}\sum_t B_t.
\end{equation}
The last inequality uses Cauchy--Schwarz. This bound allows arbitrary correlations between positions within a rollout. It also holds for independent rollouts with fixed source counts $N_q=\lambda N$ and $N_p=(1-\lambda)N$: stratifying by source removes the nonnegative between-source contribution to the variance of the iid mixture estimator.

\textbf{Asymptotic ESS gain.} Fix a position and define the overlap between the student and target prefix distributions by
\begin{equation}
\label{eq:prefix-overlap}
\kappa:=\sum_{u:m(u)>0}\frac{q(u)p(u)}{m(u)}
=\mathbb{E}_{u\sim p}[w_q(u)]
=\mathbb{E}_{u\sim q}[w_p(u)].
\end{equation}
For either component $\ell\in\{q,p\}$, $\mathbb{E}_m[w_\ell]=1$. Multiplying $m=\lambda q+(1-\lambda)p$ by $q/m$ and summing over prefixes gives $\lambda\mathbb{E}_m[w_q^2]+(1-\lambda)\kappa=1$; interchanging the components gives $(1-\lambda)\mathbb{E}_m[w_p^2]+\lambda\kappa=1$. Thus the population ESS fractions are
\begin{equation}
\label{eq:ess-population}
\begin{aligned}
\rho_q&:=\frac{(\mathbb{E}_m[w_q])^2}{\mathbb{E}_m[w_q^2]}
=\frac{\lambda}{1-(1-\lambda)\kappa}\ge\lambda,\\
\rho_p&:=\frac{(\mathbb{E}_m[w_p])^2}{\mathbb{E}_m[w_p^2]}
=\frac{1-\lambda}{1-\lambda\kappa}\ge1-\lambda.
\end{aligned}
\end{equation}
Nonnegativity gives $\kappa\ge0$, and $\mathbb{E}_m[w_q^2]\ge(\mathbb{E}_m[w_q])^2=1$ gives $\kappa\le1$. Whenever the prefix distributions overlap, $\kappa>0$, so both inequalities in \eqref{eq:ess-population} are strict. For $\lambda=1/2$, both fractions equal $1/(2-\kappa)$.

For $N$ independent prefixes $U_i\sim m$, write $w_i=w_\ell(U_i)$. The empirical statistic is $\mathrm{ESS}_N=(\sum_{i=1}^N w_i)^2/\sum_{i=1}^N w_i^2$, defined as zero if all sampled ratios vanish. Since the importance ratios are bounded, the law of large numbers gives $\mathrm{ESS}_N/N\to\rho_\ell$ almost surely. The same limits hold for fixed source counts whose proportions converge to $\lambda$ and $1-\lambda$, by applying the law of large numbers within each source.

In separate pools with these proportions, each term uses only its own source with unit importance ratios, giving ESS fractions $\lambda$ and $1-\lambda$. Sharing therefore strictly improves both asymptotic ESS fractions whenever $\kappa>0$.

\subsection{Token-level skew divergence}
\label{app:skew}
Condition on fixed $(x,y^s,F)$ and a finite horizon, and let $q$ and $p$ have common support. For two distributions $a,b$, define
\begin{equation}
K_\alpha(a\,\|\,b):=\mathrm{KL}\big(a\,\|\,(1-\alpha)b+\alpha a\big),
\qquad 0<\alpha<1.
\end{equation}
The mixture is at least $\alpha a$ pointwise, and convexity of KL in its second argument gives
\begin{equation}
\label{eq:skew-token-bound}
0\le K_\alpha(a\,\|\,b)
\le \min\!\left\{\log(1/\alpha),\,(1-\alpha)\,\mathrm{KL}(a\,\|\,b)\right\}.
\end{equation}
For a finite vocabulary and common support, continuity gives $K_\alpha(a\,\|\,b)\to\mathrm{KL}(a\,\|\,b)$ as $\alpha\to0$. Summing the token KL terms under their respective visitation laws, or equivalently using the ideal mixture and weights in \eqref{eq:balance}, yields
\begin{equation}
\label{eq:skew-jeffreys}
0\le\mathcal{L}_\alpha\le(1-\alpha)\,\mathrm{J}(q,p),
\qquad \lim_{\alpha\to0}\mathcal{L}_\alpha=\mathrm{J}(q,p),
\end{equation}
where $\mathcal{L}_\alpha$ denotes \eqref{eq:objective} with generic target $p$. Taking $p=p^{\mathrm{corr}}$ gives the method's ideal skew objective.

To distinguish skewing next-token distributions from skewing distributions over rollouts, consider the mixture $M^{\mathrm{rev}}_\alpha=\alpha q_{\mathrm{seq}}+(1-\alpha)p_{\mathrm{seq}}$. Bayes' rule gives its next-token conditional as
\begin{equation}
\label{eq:skew-posterior}
M^{\mathrm{rev}}_{\alpha,t}
=\widetilde\lambda_t q_t+(1-\widetilde\lambda_t)p_t,
\qquad
\widetilde\lambda_t
=\frac{\alpha q(y_{<t})}{\alpha q(y_{<t})+(1-\alpha)p(y_{<t})}.
\end{equation}
The chain rule then gives
\begin{equation}
\mathrm{KL}\big(q_{\mathrm{seq}}\,\|\,M^{\mathrm{rev}}_\alpha\big)
=\sum_t\mathbb{E}_{y_{<t}\sim q}\!\left[\mathrm{KL}\big(q_t\,\|\,M^{\mathrm{rev}}_{\alpha,t}\big)\right],
\end{equation}
with the forward term obtained by interchanging $q$ and $p$. Applying \eqref{eq:skew-token-bound} to the rollout distributions bounds the sum of these two rollout-level terms by $2\log(1/\alpha)$, independently of the horizon. Moreover, $M^{\mathrm{rev}}_{\alpha,t}\ge\widetilde\lambda_t q_t$ implies
\begin{equation}
0\le\mathrm{KL}\big(q_t\,\|\,M^{\mathrm{rev}}_{\alpha,t}\big)
\le-\log\widetilde\lambda_t\;\longrightarrow\;0
\quad\text{as }\frac{q(y_{<t})}{p(y_{<t})}\longrightarrow\infty.
\end{equation}
The analogous suppression holds for the forward term when the prefix increasingly favors $p$. Keeping the token coefficient at $\alpha$ removes this dependence on accumulated prefix evidence; its divergence bound applies separately at each position.

\section{Synthetic preference-reversal experiment}
\label{app:score-reversal}
\textbf{Construction and A3.} We prescribe teacher priors $b$ and posteriors $f$ over a correct token $c$, a wrong token $n$, and a third token $o$. In $(c,n,o)$ order, ordinary examples use $b=(0.40,0.01,0.59)$ and $f=(0.60,0.012,0.388)$; high-prior examples, a fraction $\eta$, use $b=(0.95,0.01,0.04)$ and $f=(0.98,0.012,0.008)$. In the latter, the complements of $c$ and $n$ retain $0.4$ and approximately $0.998$ of their prior probabilities. Under A0--A2, $S(\bar c)<S(\bar n)$; A3 therefore requires $S(c)>S(n)$. The log-probability difference instead gives $0.031<0.182$, whereas relative distance gives $0.600>0.002$.

\textbf{From scores to updates.} Both methods use $p_i(v)\propto q_{\mathrm{old}}(v\mid x_i)e^{S_i(v)}$ ($\beta=1$), where $q_{\mathrm{old}}$ is a fixed copy of the current student, and minimize mean reverse KL toward these targets. The score controls the change in target odds:
\[
\log\frac{p_i(c_i)}{p_i(n_i)}-\log\frac{q_{\mathrm{old}}(c_i\mid x_i)}{q_{\mathrm{old}}(n_i\mid x_i)}
=S_i(c_i)-S_i(n_i).
\]
Thus, on high-prior examples, the log-probability difference shifts relative preference toward the wrong token.

\textbf{Exact optimization (panel a).} A single shared token distribution starts uniform and undergoes $K=20$ exact updates. Minimizing mean reverse KL gives the normalized geometric mean of the targets, yielding $\log[q_K(c)/q_K(n)]=K\mathbb{E}_\eta[S(c)-S(n)]$. The log-probability difference reverses preference at $\eta\approx0.596$. Every strictly increasing transformation of relative distance ranks $c$ above $n$ in both example types, keeping these log-odds positive. The plotted relative-distance curves use the identity scale $S=d$.

\textbf{Neural training (panels b,c).} We fix 1,000 inputs $x_i\sim\mathcal{N}(0,I_{32})$ with independently permuted token roles per example. Across $\eta$, inputs and roles remain fixed and high-prior sets are nested, with exactly $1{,}000\eta$ examples. A $32$--$128$--$3$ MLP with GELU trains with Adam (learning rate $10^{-3}$, no weight decay, batch size 64) for 50 rounds of 25 updates. Targets are rebuilt between rounds and held fixed within each; optimizer state persists. Initialization and batch order are paired across methods for each of five seeds. Panel (b) averages $\log[q(c_i\mid x_i)/q(n_i\mid x_i)]$ over training examples; panel (c) reports three-token argmax accuracy. Curves show means $\pm$ standard errors across seeds, evaluated on this fixed training set.

\section{Practical implementation}
\label{sec:impl}

\textbf{Bounded token divergences.} To limit the magnitude of each token loss, we replace each KL term by an $\alpha$-skew divergence \citep{lee2001skew}. Define $m^{\mathrm{rev}}_t=(1-\alpha)p^{\mathrm{corr}}_t+\alpha q_t$ and $m^{\mathrm{fwd}}_t=(1-\alpha)q_t+\alpha p^{\mathrm{corr}}_t$. The ideal pooled loss becomes
\begin{equation}
\label{eq:objective}
\mathcal{L}_\alpha
=\mathbb{E}_{y\sim m_{\mathrm{seq}}}\sum_t
\Big[w_q(y_{<t})\,\mathrm{KL}(q_t\,\|\,m^{\mathrm{rev}}_t)
+w_{p^{\mathrm{corr}}}(y_{<t})\,\mathrm{KL}(p^{\mathrm{corr}}_t\,\|\,m^{\mathrm{fwd}}_t)\Big],
\end{equation}
where $m_{\mathrm{seq}}=\lambda q_{\mathrm{seq}}+(1-\lambda)p^{\mathrm{corr}}_{\mathrm{seq}}$ and the weights are given by \eqref{eq:balance}. We use $\alpha=0.01$ at every prefix. Each unweighted token divergence is at most $\log(1/\alpha)$; the ideal loss is bounded above by $(1-\alpha)\mathrm{J}(q,p^{\mathrm{corr}})$ and recovers Jeffreys divergence as $\alpha\to0$. Keeping $\alpha$ fixed at each token avoids the cap on the full-rollout loss and diminishing conditional losses that can arise from skewing distributions over rollouts. Appendix~\ref{app:skew} derives this distinction; Table~\ref{tab:skew-value-gap} compares the skew and unskewed objective values for perturbed neural policies.

\textbf{Efficient rollouts.} For each training example $(x,y^s,F)$, we reuse $y^s$ as the student rollout in the pool and generate rollouts from the feedback-conditioned teacher $\pi_F$. Because the feedback is written about $y^s$, supervising the student at the prefixes of $y^s$ places the feedback's assessment where it applies. Sampling teacher rollouts from $\pi_F$ rather than from $p^{\mathrm{corr}}$ lets the student and both teacher-branch scores be computed afterward in batched passes over the pool instead of at every decoding step; $\pi_F$ then serves as the proposal for the forward term, whose target remains $p^{\mathrm{corr}}$.

The implemented weights use the mixture denominator $m_{\mathrm{raw}}(u)=\lambda q(u)+(1-\lambda)\pi_F(u)$, evaluated with the feedback context for each rollout. The reverse weight is $q(u)/m_{\mathrm{raw}}(u)$, which keeps the pointwise bound $1/\lambda$. The forward weight is $\min\{p^{\mathrm{corr}}(u)/m_{\mathrm{raw}}(u),c_{\mathrm{IS}}\}$, clipped at $c_{\mathrm{IS}}$ because $\pi_F$ is a proposal for $p^{\mathrm{corr}}$ rather than one of its mixture components. With $\lambda=1/2$ and $c_{\mathrm{IS}}=2$, clipping touched about $1\%$ of forward weights on average in each of three EmbodiedEval training runs, and at most $4.2\%$ in any single update (Appendix~\ref{app:rollout-diagnostics}).

\textbf{Vocabulary approximation.} Both token losses sum over the full vocabulary at each supervised prefix. We obtain top-$k$ probabilities from each teacher branch and compute the score on their union $U=\mathcal{V}_k^f\cup\mathcal{V}_k^b$. A probability missing from one branch is assigned a common positive floor, and the score is set to zero outside $U$. The resulting target remains normalized over the full vocabulary: its normalizer is $Z_t=\sum_{v\in U}\bar q(v)e^{\beta S(v)}+1-\sum_{v\in U}\bar q(v)$, and its probabilities outside $U$ are $\bar q(v)/Z_t$. This form permits analytic evaluation of the remaining vocabulary contribution to each loss; the approximation is in the teacher-derived score.

\textbf{Gradient convention.} Gradients flow through the explicit occurrences of $q$ in both token divergences, including their skew mixtures. The corrected target, its normalizer, the sampled prefixes, and the importance weights are held fixed. Training therefore uses a semi-gradient of the token losses.

\textbf{Numerical comparison for perturbed neural policies.} Table~\ref{tab:skew-value-gap} reports the relative value gap $\delta_\alpha=(\mathrm{J}-\mathcal{L}_\alpha)/\mathrm{J}$, for $\mathrm{J}>0$, at $\alpha=0.01$ for neural autoregressive policy pairs. The pairs use random networks with an eight-token context; $p$ is a parameter perturbation of $q$, calibrated in $\mathrm{KL}(q_t\,\|\,p_t)$ on student prefixes.

\begin{table}[!ht]
\centering
\small
\begin{tabular}{lcr}
\hline
Policy pair & Vocabulary size, horizon & Relative gap \\
\hline
Neural, KL calibration $0.02$ nats & $32{,}768,\ 512$ & $2.0\%$ \\
Neural, KL calibration $0.5$ nats & $32{,}768,\ 512$ & $3.3\%$ \\
\hline
\end{tabular}
\caption{Relative reduction from unskewed Jeffreys for perturbed neural policies. Entries are Monte Carlo estimates using $4{,}096$ rollouts per policy, shared prefixes for both objectives, and exact vocabulary sums.}
\label{tab:skew-value-gap}
\end{table}

\section{Forward-weight clipping on EmbodiedEval}
\label{app:rollout-diagnostics}
We measured forward importance-weight clipping in three training runs on EmbodiedEval, using raw-teacher rollouts with $\lambda=1/2$ and $c_{\mathrm{IS}}=2$. The mean clipping rates were $1.11\%$, $0.96\%$, and $1.02\%$; the largest recorded rate across the three runs was $4.19\%$. Rates are computed as the fraction of valid token prefixes in the rollouts whose forward weight exceeds $2$, averaged over microbatches.

\section{Experimental details}
\label{app:exp-details}

\subsection{Benchmarks}
\label{app:benchmarks}

\textbf{Trivia Fantasy.} This synthetic benchmark contains 20 fictional facts, each stating where an object is located. A simulated user poses a 10-option multiple-choice question about each fact. Neither the student nor the teacher initially knows the answers, which exist only in the judge's answer key; all task information reaches the student through the judge's feedback. The student assigns near-zero probability to the correct option on some questions. Once its rollout commits to an incorrect option, the feedback-conditioned teacher tends to continue from that choice, leaving only the few preceding tokens to carry the correction. Evaluation measures recall of the taught facts by checking the selected option against the answer key, without feedback or a judge.

\begin{figure}[t]
\centering
\includegraphics[width=0.6\linewidth]{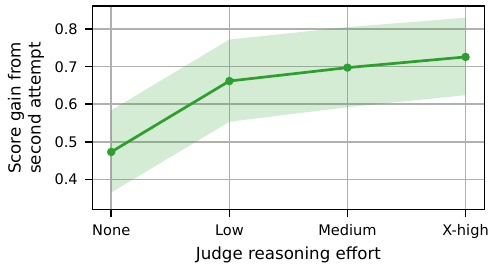}
\caption{Mean score difference between teacher rollouts generated with and without feedback on EmbodiedEval (10 repetitions). Most of the increase comes from enabling judge reasoning, with smaller gains at higher effort. Shading shows 95\% confidence intervals.}
\label{fig:channel-scaling}
\end{figure}

\textbf{EmbodiedEval.} This internal benchmark evaluates the vision-language planner of an embodied robot that moves and performs tasks in a human-centric environment. The planner acts through tool calls in three-turn conversations with a simulated user and must comply with detailed behavioral guidelines. Training conversations are drawn in equal proportion from two pools of 1,751 and 2,014 scenarios. Evaluation uses 12 held-out suites of 20 tasks each, targeting common failure cases. A task passes when the suite's judge detects no violation, and the reported score is the mean pass rate across suites. Assessing compliance requires reasoning about the task, available actions, and observations; enabling judge reasoning makes its feedback more useful for revising rollouts (Figure~\ref{fig:channel-scaling}). None of these criteria have programmatic verifiers, motivating our use of verbal feedback in this setting.

\textbf{ToolAlpaca.} ToolAlpaca~\citep{tang2023toolalpaca} pairs natural-language requests and API specifications with the required tool calls. Its original corpus was generated through simulated interactions with diverse APIs, function names, parameter schemas, and user intents. We use SDPO's static call-prediction adaptation and released split~\citep{hubotter2026reinforcement}, with 4,046 training requests and 68 held-out requests involving tools absent from training. SDPO's scorer checks the predicted function names and arguments against the reference calls.

\textbf{SciKnowEval.} We use the Level-3 scientific reasoning subsets of SciKnowEval~\citep{feng2024sciknoweval}, covering deduction, quantitative calculation, and prediction in biology, chemistry, physics, and materials science. We retain SDPO's released per-domain splits~\citep{hubotter2026reinforcement}: 450, 1,890, 720, and 841 training questions, respectively, with held-out sizes listed in Table~\ref{tab:hparams-bench}. Questions are multiple-choice, and evaluation scores the extracted final option.

\subsection{Training and implementation}
Tables~\ref{tab:hparams-bench} and~\ref{tab:hparams-method} list the settings shared by all methods on each benchmark and the settings specific to each method.

\textbf{Feedback handling.} When no correction is needed, the judge returns a fixed acknowledgement, which we omit from the teacher's context. The EmbodiedEval evaluation judge is Gemini~3.1 Pro.

\textbf{Input format.} On SciKnowEval and ToolAlpaca, the prompt is limited to 2,048 tokens and the model's thinking mode is disabled, as in SDPO.

\textbf{Baseline implementation.} For SDPO, the student retains its ordinary task context, while the teacher receives the judge's feedback in addition to the task context and sampled prefix, without a separate previous-attempt summary. Its vocabulary approximation represents the remaining probability mass with a single tail bucket. The baseline importance weights in Table~\ref{tab:hparams-method} are held fixed during differentiation; $\pi_{\mathrm{beh}}$ is the policy that generated the rollout, and W2S-OPD's anchor $\pi_0$ is the frozen initial student evaluated under the student's ordinary context. W2S-OPD's target $\widetilde p_{\mathrm{W2S}}$ reweights $\pi_0(v)$ by $[f(v)/b(v)]^{0.75}$ on the intersection of the anchor's and both teacher branches' top-20 lists, each augmented with the sampled token. We renormalize this subset to preserve its anchor mass; other retained anchor probabilities remain unchanged.

\textbf{Compute.} We implement all methods in verl~\citep{sheng2025hybridflow}, using Megatron-LM~\citep{shoeybi2019megatron} for training and vLLM~\citep{kwon2023efficient} for generation and teacher scoring, and run them on NVIDIA A100 and RTX PRO 6000 GPUs. Rollout generation runs asynchronously with training, with synchronization and reuse settings listed in Table~\ref{tab:hparams-bench}.

\begin{table}[!ht]
\centering
\small
\setlength{\tabcolsep}{3pt}
\renewcommand{\arraystretch}{1.12}
\begin{tabular}{@{}p{0.18\textwidth}p{0.18\textwidth}p{0.18\textwidth}p{0.18\textwidth}p{0.18\textwidth}@{}}
\toprule
& Trivia Fantasy & EmbodiedEval & SciKnowEval (L3) & ToolAlpaca \\
\midrule
Student & Qwen3.5-2B & Qwen3.5-4B & Qwen3.5-4B & Qwen3.5-4B \\
Training set & 20 questions & 3,765 scenarios & 450--1,890 per subject & 4,046 requests \\
Evaluation set & 20 questions & 12 suites $\times$ 20 tasks & 50 / 210 / 80 / 94 (bio. / chem. / phys. / mat.) & 68 requests \\
Learning rate & $10^{-6}$ & $10^{-6}$ & $10^{-7}$ & $5\times10^{-7}$ \\
Warmup steps & 5 & 50 & 50 & 50 \\
Sequences per update & 512 & 128 & 128 & 128 \\
Updates per weight sync & 5 & 4 & 4 & 4 \\
Rollout reuse & until stale, least-used first & until stale, least-used first & once & once \\
Generation limit & 256 tokens per turn, at most 64 thinking & 512 tokens per call, 3 user turns & 4,096 tokens, thinking off & 4,096 tokens, thinking off \\
Judge reference & answer key & behavioral criteria & correct option & reference calls \\
Judge reasoning effort & low & low & low & low \\
User simulator & Gemini 2.5 Flash-Lite & Gemini 2.5 Flash-Lite & --- & --- \\
Evaluation metric & accuracy & mean suite pass rate & avg@16 & avg@16 \\
Evaluation sampling & $T=1$, top-$p$ $0.95$ & $T=1$, top-$p$ $0.95$ & $T=0.6$, top-$p$ $0.95$, $\le$4,096 tokens & $T=0.6$, top-$p$ $0.95$, $\le$4,096 tokens \\
Evaluation interval (steps) & 10 & 10 & 20 & 20 \\
Base-model passes & 16 & 5 & 3 & 3 \\
\bottomrule
\end{tabular}
\caption{Settings shared by all methods on each benchmark. All methods use Adam with weight decay $0.01$, gradient-norm clipping at $1$, a constant learning rate after linear warmup, training sampling at temperature $1$ and top-$p$ $0.95$, and Gemini~3.7 Flash at temperature $1$ as the feedback judge. A rollout becomes stale once the policy that generated it is more than four weight synchronizations old.}
\label{tab:hparams-bench}
\end{table}

\begin{table}[!ht]
\centering
\small
\setlength{\tabcolsep}{3pt}
\renewcommand{\arraystretch}{1.12}
\begin{tabular}{@{}p{0.17\textwidth}p{0.25\textwidth}p{0.23\textwidth}p{0.27\textwidth}@{}}
\toprule
& Ours & SDPO & W2S-OPD \\
\midrule
Teacher & frozen initial student & frozen initial student & frozen initial student \\
Teacher contexts & with and without feedback & with feedback & with and without feedback \\
Training rollouts & student and teacher, $\lambda=1/2$ & student only & student only \\
Token divergence & $\alpha$-skew Jeffreys, $\alpha=0.01$ & reverse KL & reverse KL \\
Target & corrected target \eqref{eq:iwkl}, $\beta=10$ & feedback-conditioned teacher & $\widetilde p_{\mathrm{W2S}}$, contrast strength $0.75$, anchor $\pi_0$ \\
Importance weights & prefix weights \eqref{eq:balance}, forward weight clipped at $c_{\mathrm{IS}}=2$ & per token, $\min\{2,\pi_\theta/\pi_{\mathrm{beh}}\}$ & per token, $\min\{2,\pi_\theta/\pi_{\mathrm{beh}}\}$ \\
\midrule
\multicolumn{4}{@{}p{0.95\textwidth}@{}}{Shared by all three: top-20 predictions from each teacher query, augmented with the sampled token; the student's top-20 tokens from each tensor-parallel vocabulary partition added to the support; log-probability floor $-14$ for tokens missing from a teacher's list; no entropy bonus; per-token losses summed over response tokens and divided by the microbatch size times the maximum response length.} \\
\bottomrule
\end{tabular}
\caption{Method-specific settings.}
\label{tab:hparams-method}
\end{table}

\end{document}